# Computational Analyses of Arabic Morphology

George Anton Kiraz*

COMPUTER LABORATORY, UNIVERSITY OF CAMBRIDGE

(St John's College)

E-mail. `George.Kiraz@cl.cam.ac.uk`

July 25, 1994

**Abstract**

This paper demonstrates how a (multi-tape) two-level formalism can be used to write two-level grammars for Arabic non-linear morphology using a high level, but computationally tractable, notation. Three illustrative grammars are provided based on CV-, moraic- and affixational analyses. These are complemented by a proposal for handling the hitherto computationally untreated problem of the broken plural.

It will be shown that the best grammars for describing Arabic non-linear morphology are moraic in the case of templatic stems, and affixational in the case of a-templatic stems. The paper will demonstrate how the broken plural can be derived under two-level theory via the 'implicit' derivation of the singular.

## 1 INTRODUCTION

Arabic is known amongst computational linguists, in particular computational morphologists, for its highly inflexional morphology. Its root-and-pattern phenomenon has become the prototype for the evaluation of the few non-linear morphological models which have emerged in recent years.[1] These models use the CV analysis of Arabic (McCarthy 1981) to illustrate how Arabic morphology can be analysed computationally. Although McCarthy's original work has been superseded by moraic and affixational analyses (McCarthy and Prince 1990b; McCarthy 1993), neither these accounts nor the challenging problem of the Arabic broken plural (McCarthy and Prince 1990a), to my knowledge, have been dealt with computationally. My debt to the findings of John McCarthy and Alan Prince will become apparent throughout my presentation; everything I have to say describing linguistic models is based on their work.

The purpose of the current paper is to present illustrative computational grammars for Arabic stems. The grammars are based on (multi-tape) two-level morphology and follow three linguistic analyses: CV, moraic and affixational. Despite the CV analysis being superseded by the other two models, it is discussed here for a number of reasons: firstly, the CV analysis serves as a good introduction to Arabic non-linear morphology and to McCarthy's original findings; secondly, it facilitates the comparison of the current work with other computational proposals which adopt the CV analysis as basis; more importantly, the non-linguist user writing a grammar of Arabic morphology may prefer to base the grammar on the CV model due to its ease of implementation and accessibility in grammatical literature.

*Supported by a Benefactor Studentship from St John's College. This research was done under the supervision of Dr Stephen G. Pulman (University of Cambridge, and SRI International) whom I thank for guidance and support. I would like to thank Dr John McCarthy for answering many of my questions. Thanks are also due to Dr John Carroll (University of Cambridge) and Ruth Glassock (St John's College) for editorial comments on an earlier draft.

[1]Non-linear proposals include Kay (1987), Kornai (1991), Wiebe (1992), Narayanan and Hashem (1993) and Bird and Ellison (1994). Beesley (1991) describes a two-level model with 'detours'.

The sample computational grammars presented here are by no means exhaustive, nor do they 'simulate' the various linguistic models proposed by McCarthy and Prince. The grammars, however, aim at showing how such linguistic models can be catered for in two-level theory. In addition, the grammars and lexica demonstrate that the chosen computational formalism is capable of addressing the problems of Arabic morphology.

The following convention has been adopted throughout the paper. Morphemes are represented in braces, { }, graphemes in square brackets, [ ], and surface forms in solidi, / /. In listings of grammars and lexica, capital letters indicate variables.

The structure of the paper is as follows: Section 2 introduces two-level theory. Sections 3, 4 and 5 discuss CV-, moraic- and affixational morphology, respectively; each section starts by describing the linguistic model in question (based on the works of McCarthy and Prince) followed by a computational account. Section 6 takes a look at the problems of the broken plural. Finally, section 7 gives concluding remarks.

## 2 THEORETICAL BACKGROUND

In the past decade, two-level morphology, introduced by Koskenniemi (1983), has become the standard morphological model in computational linguistics. Motivated by the highly inflexional, but linear, morphology of Finnish, the model has been applied successfully to various languages. Two-level morphology assumes two levels of representation, lexical and surface, with the mapping between them implemented as finite-state transducers (FSTs). Further, it assumes that surface forms, say /moved/, are described by the *concatenation* of lexical morphemes, e.g. {move} + {ed}. This assumption makes it very difficult to adapt the model to non-linear morphology. For example, Arabic /kutib/ 'to write - perfect passive' consists of the morphemes {ktb} 'notion of writing' and {ui} 'perfect passive', whose concatenation results in the ill-formed */ktbui/. To overcome this limitation, Kay (1987) proposed a model where lexical morphemes occupy multiple tapes. Motivated by Kay's work and a two-level formalism reported by Pulman and Hepple (1993), Kiraz (1994) proposed a multi-tape two-level model accompanied by a multi-tape two-level formalism.

Section 2.1 introduces two-level theory; section 2.2 takes a brief look at the development in two-level formalisms which is of relevance to this work; finally, section 2.3 defines multi-tape two-level morphology.

### 2.1 The Two-Level Model

Two-level theory defines two levels of strings in recognition and synthesis: lexical strings represent morphemes, and surface strings represent surface forms. Two-level morphological rules map the two strings; the rules are compiled into FSTs. An abstract example of a two-level rule appears in (1).

(1) $a : b \quad \Rightarrow \quad c : d \_\_\_ e : f$

The rule states that lexical *a* maps to surface *b* when preceeded by lexical *c* mapping to surface *d* and followed by lexical *e* mapping to surface *f*. An example showing the derivation of /moved/ is given in (2) (Ritchie 1992).

(2) DERIVATION OF /moved/

| m | o | v | e | + | e | d | *lexical tape* |
| 1 | 1 | 1 | 3 | 2 | 1 | 1 |
| m | o | v | 0 | 0 | e | d | *surface tape* |

Rules:
1. Identity Rule: X:X $\quad \Rightarrow \quad \_\_\_$
2. Boundary Rule: +:0 $\quad \Rightarrow \quad \_\_\_$
3. [e]-deletion Rule: e:0 $\quad \Rightarrow \quad$ v:v $\_\_\_$ +:0

(2) shows two strings (tapes): the lexical string is a concatenation of the two lexical morphemes {move} and {ed}, separated by the lexical boundary symbol '+'. The surface string represents the surface form /moved/ (after deleting the null symbols, '0'). Mapping between the two strings is defined by the rules in (2). The numbers between the two strings indicate the rules which sanction the moves. The identity rule allows a lexical character to appear on the surface, e.g. lexical [m] maps to surface [m]. The boundary rule allows the morpheme boundary symbol '+' to be deleted on the surface, i.e. '+' surfaces as '0'. The [e]-deletion rule states the deletion of lexical [e] in {move} in the context shown. The above example demonstrates that the lexical string in standard two-level morphology represents the concatenation of lexical morphemes. It is not possible to derive an Arabic surface form such as /kutib/ by the concatenation of its constituent morphemes {ktb} and {ui}.

## 2.2 Development in Two-Level Formalisms

It was pointed out by Black et al. (1987) that previous two-level rules, cf. (2), affect one character at a time and proposed a formalism which maps between (equal numbered) sequences of surface and lexical characters using the formalism in (3).

(3) BLACK *et al.*'s FORMALISM
   SURF $\Leftrightarrow$ LEX

A lexical string maps to a surface string iff they can be partitioned into pairs of lexical-surface subsequences, where each pair is licenced by a rule. Ruessink (1989) added explicit contexts and allowed unequal sequences. Pulman and Hepple (1993) allowed feature-based representations interpreted via unification; this allows a lexical entry to include variables, e.g. the Arabic verbal prefix {tV}, where the quality of V is determined from the vowel of the following stem as in /ta+kaatab/ and /tu+kuutib/. The Pulman-Hepple/Ruessink/Black *et al.* formalism is given in (4).

(4) THE PULMAN-HEPPLE/RUESSINK/BLACK *et al.* FORMALISM
   a. LSC - SURF - RSC $\Rightarrow$ LLC - LEX - RLC
   b. LSC - SURF - RSC $\Leftrightarrow$ LLC - LEX - RLC

   where
   | | | | | |
   |---|---|---|---|---|
   | LSC | = | left surface context | LLC | = left lexical context |
   | SURF | = | surface form | LEX | = lexical form |
   | RSC | = | right surface context | RLC | = right lexical context |

The special symbol * indicates an empty context, which is always satisfied. The operator $\Rightarrow$ states that LEX *may* surface as SURF in the given context, while the operator $\Leftrightarrow$ adds the condition that when LEX appears in the given context, then the surface description *must* satisfy SURF. The latter caters for obligatory rules. The advantage of this formalism over others is that it allows *inter alia* mappings between lexical and surface strings of unequal lengths. This advantage will become apparent in the rest of this paper. To demonstrate this formalism, the rules in (2) are rewritten in (5).

(5) TWO-LEVEL EXAMPLES
   a. Default Rule: * - X - *   $\Rightarrow$   * - X - *
   b. Boundary Rule: * -   - *   $\Rightarrow$   * - + - *
   c. [e]-deletion Rule: * -   - *   $\Leftrightarrow$   v - e - +

Note that the blank character '0' is indicated here by blank in SURF.

Another advantage of the above formalism is rule features. Rules are associated with a feature structure which must unify with the lexical feature structure of the morpheme affected by the rule. This allows a rule to be applied on a specific category of morphemes.

## 2.3 Multi-Tape Two-Level Morphology

The notion of using multiple tapes to represent autonomous morphemes first appeared in Kay (1987). Motivated by Kay's work, Kiraz (1994) proposed a multi-tape two-level model which can account for Semitic root-and-pattern morphology using high level notation. The model adopts the Pulman-Hepple/Ruessink/Black *et al.* formalism with two extensions. The first extension is that all expressions in the lexical side of the rules (i.e. LLC, LEX and RLC) are $n$-tuple of regular expressions of the form in (6).

(6) ( $x_1$, $x_2$, ..., $x_n$ )

The $i$th expression refers to symbols on the $i$th tape. When $n = 1$, the parentheses can be ignored; hence, $(x)$ and $x$ are equivalent. Since $n$-tape two-level rules affect $n$ lexical tapes, rules are associated with $n$ feature structures of the form in (7).

(7) ([$a_{11}$=$v_{11}$, $a_{12}$=$v_{12}$, ...], [$a_{21}$=$v_{21}$, $a_{22}$=$v_{22}$, ...], ..., [$a_{n1}$=$v_{n1}$, $a_{n2}$=$v_{n2}$, ...])

Each feature structure consists of an unordered set of `attribute=value` pairs. The $i$th feature structure must unify with the lexical feature structure of the morpheme affected by the rule on the $i$th tape. Similar to lexical expressions, when $n = 1$, the parentheses can be ignored.

The second extension is giving LLC the ability to contain ellipsis, ..., which indicates the (optional) omission from LLC of tuples, provided that the tuples to the left of ... are the first to appear on the left of LEX. For example, the LLC expression in (8a) matches all the regular expressions in (8b).

(8) a. ( a ) ··· ( b )
    b. ab, $ax_1$b, $ax_1x_2$b, $ax_1x_2$...b, where $x_i \neq$ (a)

To illustrate how the formalism works, assume a morphological grammar which requires three lexical tapes. The first tape reads consonants, the second reads vowels and the third reads digits. Four abstract rules are given in (9).

(9) MULTI-TAPE TWO-LEVEL ABSTRACT RULES
   a. * - CDV - *   ⇒   * - (C,V,D) - *    where C ≠ t.
   b. * - tDV - *   ⇒   * - (t,V,D) - *
   c. * - - *       ⇒   (*,*,X) - ( , ,D) - (*,*,X)
   d. * - DnV - *   ⇒   (*,V,D) ··· - ( ,u, ) - (*,V,*)
   where C is a consonant, V is a vowel and D is a digit.

Rule (9a) states that reading a consonant C from the first lexical tape, a vowel V from the second tape and a digit D from the third tape, i.e. LEX = (C,V,D), corresponds to CDV on the surface tape. Rule (9b) is the same as rule (9a) except that the symbol read from the first tape must be [t].

Rule (9c) states that if (i) a digit D is read from the third tape, without reading any symbol from the first or second tapes, i.e. LEX = ( , ,D), and (ii) another digit X appears to the left and right of LEX, regardless of what appears on the first two tapes, i.e. LLC = RLC = (*,*,X); then the current digit is deleted on the surface.

Rule (9d) is more complicated: it states that if (i) [u] is read from the second tape, without reading any symbol from the first or third tapes, i.e. LEX = ( ,u, ), (ii) previously, a vowel V was read from the second tape and a digit D from the third tape, regardless of what was read from the first tape, i.e. LLC = (*,V,D) ···, and (iii) the same V appears to the right of LEX, regardless of what appears on the first and third tapes, i.e. RLC = (*,V,*); then, (i) [u] is deleted on the surface and (ii) the previously read D appears on the surface, followed by an inserted [n], followed by the previously read V. The rules are illustrated in (10).

(10) MULTI-TAPE ABSTRACT EXAMPLE

| 2 | 1 | 2 | | 7 | *digits tape* |
|---|---|---|---|---|---|
| i | | o | u | i | *vowels tape* |
| k | | t | | f | *consonants tape* |
| a | c | b | d | a | |
| k2i | | t2o2ni | f7i | | *surface tape* |

The letters between the surface tape and the lexical tapes refer to the rules in (9) which sanction the moves. The example in (10) maps the surface string /k2it2o2nif7i/ to the lexical strings {ktf}, {ioui} and {2127}.

The implementation used here consists of two modules (Trost 1990): a (multi-tape) two-level system which handles non-linear morphology, and a shift-reduce parser which handles morphotactics using unification-based grammars. A surface string corresponds to a lexical string iff the two strings are passed successfully through the two-level module, and the lexical sequences produced are parsed successfully by the parser.

The remaining sections of this paper demonstrate how the above formalism can be applied to Arabic stems following various linguistic models, specifically CV-, moraic- and inflexional based grammars.

## 3 CV-BASED ANALYSIS

The CV analysis presented here describes the Arabic verbal system. The basic stems of the perfect appear in (11) (McCarthy 1981).[2]

(11) ARABIC VERBAL STEMS

| | Active | Passive | | Active | Passive | | Active | Passive |
|---|---|---|---|---|---|---|---|---|
| 1 | katab | kutib | 8 | ktatab | ktutib | 15 | ktanbay | |
| 2 | kattab | kuttib | 9 | ktabab | | Q1 | daħraĵ | duħriĵ |
| 3 | kaatab | kuutib | 10 | staktab | stuktib | Q2 | tadaħraĵ | tuduħriĵ |
| 4 | ʔaktab | ʔuktib | 11 | ktaabab | | Q3 | dħanraĵ | dħunriĵ |
| 5 | takattab | tukuttib | 12 | ktawtab | | Q4 | dħarĵaĵ | dħurĵiĵ |
| 6 | takaatab | tukuutib | 13 | ktawwab | | | | |
| 7 | nkatab | nkutib | 14 | ktanbab | | | | |

The quality of the second vowel in Measure 1 is different from one root to another; it is lexically marked, e.g. /qatal/ 'to kill', /samuħ/ 'to be generous', /samiʕ/ 'to hear', from the roots {qtl}, {smħ} and {smʕ}, respectively.

Section 3.1 presents McCarthy's CV analysis, and section 3.2 presents the corresponding computational analysis. Since the derivation of the verbs in section 3.1 are not in the order presented in (11), measure numbers appear in the margin for ease of reference.

### 3.1 Linguistic Description

McCarthy (1981) proposed a linguistic model for Arabic morphology under the framework of autosegmental phonology (Goldsmith 1976) where a stem is represented by three types of morphemes: root morphemes consist of consonants, vocalism morphemes consist of vowels, and pattern morphemes are CV-skeleta (i.e. strings of Cs and Vs); some stems include affix morphemes, e.g. {st} in /staktab/ (Measure 10). Each morpheme sits on its own autonomous tier in the autosegmental model; the morphemes are coordinated with association lines according to the principles of autosegmental phonology. The analysis of /katab/ (Measure 1) produces three morphemes: the root   M 1 morpheme {ktb} 'notion of writing', the vocalism morpheme {a} 'perfect - active', and the pattern morpheme {CVCVC} 'Measure 1'.

---

[2]As indicated by McCarthy (1981), the data provides stems in underlying morphological forms. Hence, it should be noted that: mood, case, gender and number marking is not shown; many stems experience phonological processing to give surface forms, e.g. /nkatab/ → /ʔinkatab/ (Measure 7); the root morphemes shown are not cited in the literature in all measures, e.g. there is no such verb as */takattab/ (Measure 5), but there is /takassab/ from the root morpheme {ksb}. Some measures do not occur in the passive.

(12) DERIVATION OF MEASURE 1

/katab/ = 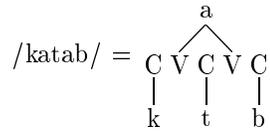

(12) illustrates how morphemes are coordinated under the principles of autosegmental phonology based on two stipulations. The first stipulation is the Well-Formedness Condition set forth in (13).

(13) WELL-FORMEDNESS CONDITION
 a. Every CV skeletal slot must be associated with at least one melody element; every consonantal melody element must be associated with at least one C slot and every vocalic melody element must be associated with at least one V slot.
 b. Association lines must not cross.

The second stipulation is the Association Convention set out in (14).

(14) ASSOCIATION CONVENTION
 Only the rightmost member of a tier can be associated to more than one member of another tier.

The association of the rightmost member of a tier to more than one member of another tier is called spreading, e.g. the spreading of [a] to two Vs in (12). The analysis of measures without affixes (i.e. Measures 3, 9 and 11) is straightforward as shown in (15).   M 3, 9, 11

(15) DERIVATION OF MEASURES 3, 9 AND 11
 Measure 3   Measure 9   Measure 11

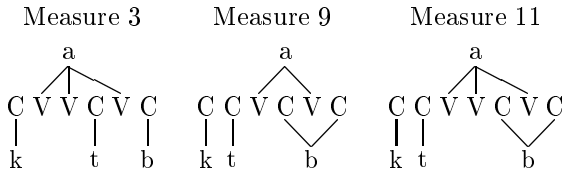

Measures with affixes are derived in a similar manner, but with associating the affix material first. The derivation of Measures 4, 6-7 and 10 is given in (16). Note that affix morphemes sit on an autonomous tier.   M 4, 6-7, 10

(16) DERIVATION OF MEASURES 4, 6-7 AND 10
 Measure 4   Measure 6   Measure 7   Measure 10

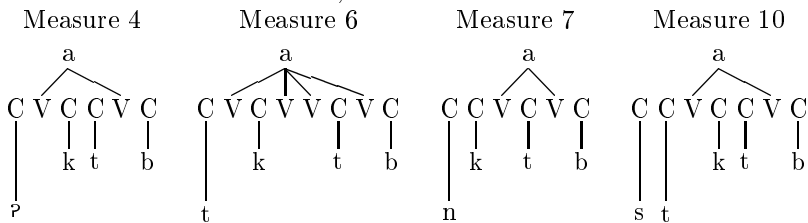

The same procedure applies to Measures 14-15, but with per-linking affixes in the initial configuration as illustrated in (17).   M 14-15

(17) DERIVATION OF MEASURES 14-15

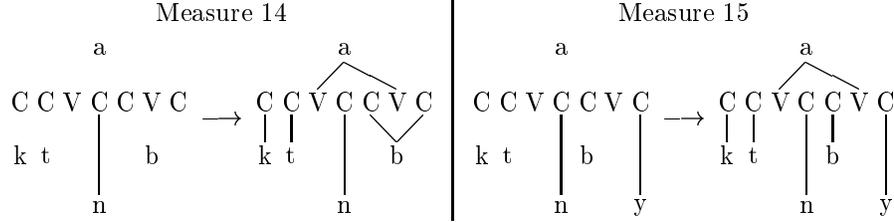

The association convention fails to produce Measures 2, 5, 8, 12 and 13. The analyses of Measures 2 and 5 require the special erasure rule in (18).  M 2, 5

(18) ERASURE RULE (MEASURES 2, 5, 12 AND 13):

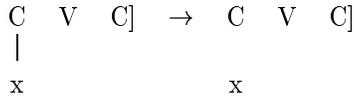

The rule deletes the link between the penultimate C and a melodic element. Applying this rule to the derivation of Measures 2 and 5 is illustrated in (19).

(19) DERIVATION OF MEASURES 2 AND 5

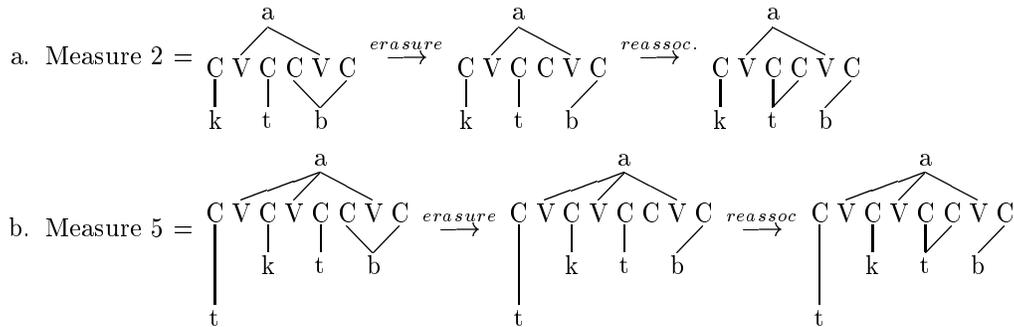

After conventional association, the forms */katbab/ and */takatbab/ are produced. To fix the error, the erasure rule is invoked. The rule deletes the association line between the penultimate C slot and [b]. Finally, the unlinked C slot is reassociated with the nearest consonant slot on the left. The same erasure rule applies to Measures 12 and 13 as shown in (20).[3]  M 12-13

(20) DERIVATION OF MEASURES 12-13

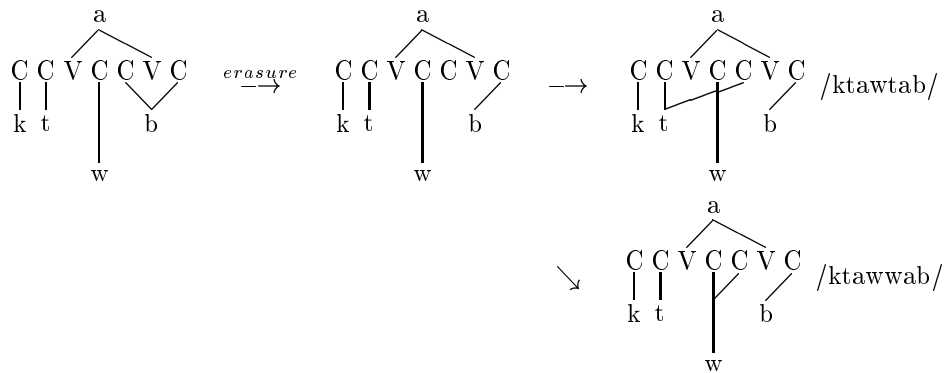

The only difference between Measures 12 and 13 is that while the second radical spreads after erasure in the former, the affix spreads in the later.

---

[3]In /ktawtab/, since {ktb} and {w} are on different tiers, the two lines linking the third C slot with [w] and the fourth C slot with [t] do not cross. Imagine the graph in three dimensions where one of the lines is on top of the other.

The derivation of Measure 8 requires a flopping rule which unlinks the reflexive {t} infix from the first C slot and links it to the second. The rule is stated in (21).

M 8

(21) FLOPPING RULE (MEASURE 8)

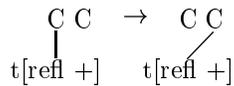

The derivation of Measure 8 is illustrated in (22).

(22) DERIVATION OF MEASURE 8

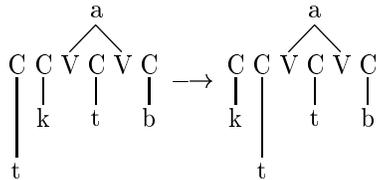

Quadrilateral measures do not pose any complications. Their derivation is shown in (23). Note that the affix in Measure Q3 is pre-linked to the template.

M Q1-Q4

(23) DERIVATION OF QUADRILATERAL MEASURES
      Measure Q1      Measure Q2      Measure Q3      Measure Q4

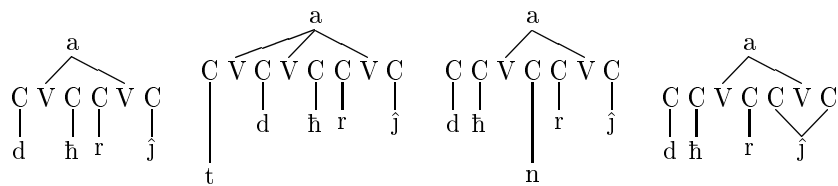

To summarise: each morpheme in the autosegmental representation lies on a separate tier. Association of morphemes follows the Well-Formedness Condition and the Association Convention. The association procedure appears in (24).

(24) CV ASSOCIATION PROCEDURE
    a. Link affixes.
    b. Link melodic elements (i.e. consonants and vowels).
    c. Apply language-specific rules; if a skeletal slot is not linked after a rule is applied, re-associate.

Measures 1, 3-4, 6-7, 9-11 are satisfied by applying steps (24a-b). Measures 12-15 have the affixes pre-associated to the CV skeleton; association then follows (24a-b), followed by applying the erasure rule to Measures 12-13. Measures 2 and 5 require the erasure rule, and Measure 8 requires the flopping rule.

## 3.2 Computational Analysis

There are two complications in the above CV analysis. Firstly, affix morphemes in Measures 12-15 must be pre-associated to the skeleton morpheme. Secondly, two language specific rules must apply on Measures 2, 5, 8, 12 and 13 in order to derive the correct form. In addition, one needs to remember that affix material is linked to the skeleton before melodic slots. Though the above analysis is linguistically justified, it can be simplified (for computational purposes) by (i) pre-compiling affixes to the template, and (ii) indexing CV slots. For example, the template CVCCVC, which describes Measures 2, 4 and Q1, becomes $c_1v_1c_2c_2v_1c_3$ for Measure 2, $ʔv_1c_1c_2v_1c_3$ for Measure 4, and $c_1v_1c_2c_3v_1c_4$ for Measure Q1. Hence, language specific rules need not be coded. Note that the Vs also need to be indexed to make sure that only the first vowel is spread in /kuutib/ (Measure 3, passive), avoiding */kuitib/.

Section 3.2.1 presents a sample lexical module, and section 3.2.2 gives a morphological grammar describing the data cited above.

### 3.2.1 The Lexica

The multi-tape two-level analysis assumes three lexical tapes: pattern tape, root tape and vocalism tape. Morphemes which fall out of the domain of root-and-pattern morphology, e.g. prefixes and suffixes, are placed on the first tape (i.e. pattern tape).[4] The lexical component is implemented as character trees (Ritchie et al. 1992).[5] Each lexical tape reads a tree which may incorporate more than one lexicon. For example, the first tree, corresponding to the first lexical tape, maintains pattern morphemes in addition to prefixes and suffixes (and may include any other morphemes not related to the stem). If a particular lexicon does not specify a tree, its entries are placed on the first tree by default. Sample entries appear in (25).

(25) CV Lexica
Number of Trees: 3.
a. Pattern Lexicon (on Tree 1):

1   $\{c_1v_1c_2v_1c_3\}$    pattern:[measure=1, tense=perf, voice=act, perf_vowel=a]
2   $\{c_1v_1c_2v_2c_3\}$    pattern:[measure=1, tense=perf, voice=act, perf_vowel=(u,i)]
3   $\{c_1v_1c_2c_2v_1c_3\}$    pattern:[measure=2, tense=perf, voice=act]
4   $\{c_1v_1v_1c_2v_1c_3\}$    pattern:[measure=3, tense=perf, voice=act]
5   $\{\textipa{P}v_1c_1c_2v_1c_3\}$    pattern:[measure=4, tense=perf, voice=act]
6   $\{tv_1c_1v_1c_2c_2v_1c_3\}$    pattern:[measure=5, tense=perf, voice=act]
    $\vdots$
7   $\{c_1c_2v_1c_3c_4v_1c_4\}$    pattern:[measure=q4, tense=perf, voice=act]
8   $\{c_1v_1c_2v_2c_3\}$    pattern:[measure=1, tense=perf, voice=pass]
9   $\{c_1v_1c_2c_2v_2c_3\}$    pattern:[measure=2, tense=perf, voice=pass]
    $\vdots$
10   $\{c_1c_2v_1c_3c_4v_2c_4\}$    pattern:[measure=q4, tense=perf, voice=pass]

b. Root Lexicon (on Tree 2):
1   {ktb} 'to write'    root:[measure=(1,2,3,4,6,7,8,10), perf_vowel=a]
2   {smħ} 'be generous'    root:[measure=(1,2,3,6,10), perf_vowel=u]
3   {smʕ} 'to hear'    root:[measure=(1,2,4,5,6,8), perf_vowel=i]

c. Vocalism Lexicon (on Tree 3):
1   {a}    vocalism:[tense=perf, voice=act, perf_vowel=a]
2   {au}    vocalism:[measure=1, tense=perf, voice=act, perf_vowel=u]
3   {ai}    vocalism:[measure=1, tense=perf, voice=act, perf_vowel=i]
4   {ui}    vocalism:[tense=perf, voice=pass]

d. Verbal Inflexional Markers (on Tree 1):
1   {a}    vim:[number=sing, person=3, gender=masc]
2   {at}    vim:[number=sing, person=3, gender=fem]

Each lexical entry is associated with a category and a feature structure (FS) of the form `cat:FS`. Feature values between parenthesis are disjunctive and are implemented using boolean vectors (Mellish (1988), Pulman (1994) provides a thorough description).

The pattern lexicon in (25a) gives CV-skeleta. Note that since CV slots are indexed, each measure requires a pattern entry for the perfect active ($25a_{1-7}$) and another for the perfect passive ($25a_{8-10}$). Measure 1, in the perfect active, requires more than one pattern entry to cater for various vocalism classes, e.g. /katab/, /samuħ/ and /samiʕ/ ($25a_{1-2}$). The root lexicon gives, for each root, the measures in which it is cited in the literature based on Wehr (1971), and indicates the perfective vowel. The vocalism lexicon provides the various vocalisms. The verbal inflexional markers lexicon provides two suffixes which are placed on the first tree.

---

[4] Note that a lexical tape in two-level morphology contains various classes of morphemes.

[5] Character trees are sequences of symbols defined in a *trie* structure (Knuth 1973).

### 3.2.2 The Grammar

Since the lexicon declares three tapes (trees), each lexical expression in the grammar must be a 3-tuple, where the $i$th element describes characters on the $i$th tape. The multi-tape two-level grammar which describes the Arabic verbal stems appears in (26).

(26) CV GRAMMAR
  a. General Rules:
   R1  Identity:    * - X - *  ⇒  * - X - *
   R2  Consonants:  * - X - *  ⇒  * - (C, X, ) - *      where C ∈ {$c_1$, $c_2$, $c_3$, $c_4$}
   R3  Vowels:      * - X - *  ⇒  * - (V, , X) - *      where V ∈ {$v_1$, $v_2$}
  b. Boundary Rules:
   R4  Non-stem:    * - - *    ⇒  (X, , ) - + - *       where X ≠ +
   R5  Stem:        * - - *    ⇒  (X,*,*) - (+,+,+) - * where X ≠ +
  c. Spreading Rules:
   R6  Radicals:    * - X - *  ⇒  (C, X, ) ··· - C - *  where C ∈ {$c_2$, $c_3$, $c_4$}
   R7  1st vowel:   * - X - *  ⇒  ($v_1$, , X) ··· - $v_1$ - *

The grammar contains three general rules given in (26a): R1 allows any character on the first lexical tape to surface, e.g. infixes, prefixes and suffixes. R2 states that any C on the first (pattern) tape and X on the second (root) tape with no transition on the third (vocalism) tape corresponds to X on the surface tape; this rule sanctions consonants. Similarly, R3 states that any V on the pattern tape and X on vocalism tape with no transition on the root tape corresponds to X on the surface tape; this rule sanctions vowels.

(26b) gives two boundary rules: R4 is used for non-stem morphemes, e.g. prefixes and suffixes. R5 applies to stem morphemes reading three boundary symbols simultaneously; this marks the end of a stem. Notice that LLC ensures that the right boundary rule is invoked at the right time.

Before embarking on the rest of the rules, an illustrated example seems in order. The derivation of /dḥunriĵa/ (Measure Q3, passive), from the three morphemes {$c_1c_2v_1nc_3v_2c_4$}, {dḥrĵ} and {ui}, and the suffix {a} '3rd person' is illustrated in (27).

(27) TWO-LEVEL DERIVATION OF MEASURE Q3 + {a}

| | | u | | | i | | + | | vocalism tape (VT) |
|---|---|---|---|---|---|---|---|---|---|
| d | ḥ | | | r | | ĵ | + | | root tape (RT) |
| $c_1$ | $c_2$ | $v_1$ | n | $c_3$ | $v_2$ | $c_4$ | + | a | + | pattern tape (PT) |
| 2 | 2 | 3 | 1 | 2 | 3 | 2 | 5 | 1 | 4 | |
| d | ḥ | u | n | r | i | ĵ | | a | | surface tape (ST) |

The numbers between the surface tape and the lexical tapes indicate the rules in (26) which sanction the moves.

Resuming the description of the grammar, (26c) presents spreading rules. Notice the use of ellipsis to indicate that there can be tuples separating LEX and LLC, as far as the tuples in LLC are the nearest ones to LEX. R6 sanctions the spreading (and gemination) of consonants. R7 sanctions the spreading of the first vowel. Spreading examples appear in (28).

(28) TWO-LEVEL DERIVATION OF MEASURES 1-3

Measure 1

| | a | | | | + | VT |
|---|---|---|---|---|---|---|
| k | | t | | b | + | RT |
| $c_1$ | $v_1$ | $c_2$ | $v_1$ | $c_3$ | + | PT |
| 2 | 3 | 2 | 7 | 2 | 5 | |
| k | a | t | a | b | | ST |

Measure 2

| | a | | | | | + | VT |
|---|---|---|---|---|---|---|---|
| k | | t | | | b | + | RT |
| $c_1$ | $v_1$ | $c_2$ | $c_2$ | $v_1$ | $c_3$ | + | PT |
| 2 | 3 | 2 | 6 | 7 | 2 | 5 | |
| k | a | t | t | a | b | | ST |

Measure 3

| | a | | | | | + | VT |
|---|---|---|---|---|---|---|---|
| k | | | t | | b | + | RT |
| $c_1$ | $v_1$ | $v_1$ | $c_2$ | $v_1$ | $c_3$ | + | PT |
| 2 | 3 | 7 | 2 | 7 | 2 | 5 | |
| k | a | a | t | a | b | | ST |

The unification-based grammar appears in (29). It consists of two rules which derive verbs. In cases when the two-level module produces various analyses, only the ones sanctioned by the grammar are considered well-formed.

(29) CV MORPHOSYNTACTIC GRAMMAR
    verb:[measure=M, tense=T, voice=V, number=N, person=P, gender=G] $\longrightarrow$
        verb_stem:[measure=M, tense=T, voice=V]
        vim:[number=N, person=P, gender=G].
    verb_stem:[measure=M, tense=T, voice=V] $\longrightarrow$
        pattern:[measure=M, tense=T, voice=V]
        root:[measure=M, perf_vowel=PV]
        vocalism:[tense=T, voice=V, perf_vowel=PV].

To summarise: The lexica include a pattern lexicon (with affixes pre complied), a root lexicon and a vocalism lexicon, each on a unique lexical tape; other affix lexica, including those which deal with non-templatic words, are placed on the first lexical tape. The grammar consists of three groups of rules: The first handles default mappings between lexical and surface tapes (R1-R3), the second deals with boundary symbols (R4-R5), and the third consists of spreading and gemination rules which make use of ellipsis in LLC (R6-R7).

# 4 MORAIC ANALYSIS

The moraic analysis described in this section is applied to the Arabic noun. Nominal stems are listed in (30) (McCarthy 1993).[6]

(30) ARABIC NOMINAL STEMS

| | Pattern | Noun | Gloss | | Pattern | Noun | Gloss |
|---|---|---|---|---|---|---|---|
| a. | CVCC | nafs | 'soul' | e. | CVVCVVC, | ĵaamuus | 'buffalo' |
| b. | CVCVC, | ʔasad | 'lion' | f. | CVCCVC, | ĵundub | 'locust' |
| c. | CVCVVC, | waziir | 'minister' | g. | CVCCVVC | ĵumhuur | 'multitude' |
| d. | CVVCVC, | kaatib | 'writer' | | | | |

Other nouns may appear with different lexically marked vocalic melodies. A linguistic description of the moraic analysis is given in section 4.1, followed by a computational account in section 4.2.

## 4.1 Linguistic Description

In McCarthy's original account (cf. § 3.1), the pattern morpheme was described as a CV-skeleton. McCarthy and Prince (1990b) argued that a different vocabulary be used to represent the pattern morpheme according to the Prosodic Morphology Hypothesis set forth in (31).

(31) PROSODIC MORPHOLOGY HYPOTHESIS
    Templates are defined in terms of the authentic units of prosody.

Moraic theory states that the phonological word is made up of feet; the foot is composed of at least one stressed syllable and may have unstressed syllables, and the syllable weight is measured by the unit mora. The prosodic hierarchy is given in (32) from top to bottom.

(32) PROSODIC HIERARCHY
    a. Phonological Word: W
    b. Foot: F
    c. Syllable: $\sigma$
    d. Mora: $\mu$

---

[6]I have replaced some entries from McCarthy and Prince (1990a) to match the plural data in section 6.

A monomoraic syllable, $\sigma_\mu$, is light (L), and a bimoraic syllable, $\sigma_{\mu\mu}$, is heavy (H). Arabic syllables are three kinds: open light, CV, open heavy, CVV, and closed heavy, CVC. This typology is represented in (33).

(33) SYLLABIC TYPOLOGY
    a. Light CV    b. Heavy CVV    c. Heavy CVC

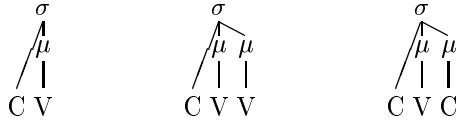

Association of Cs and Vs to templates takes the following form: a node $\sigma$ always takes a C, and a mora $\mu$ takes a V, e.g. (33a); however, in bimoraic syllables, the second $\mu$ may be associated to either a C or a V, e.g. (33b-c).[7] Hence, $\sigma_{\mu\mu}$ does not distinguish between CVV and CVC syllables.

The moraic analysis builds on the notion of extrametricality. At the right edge of stems, all Arabic stems - nominal and verbal - must end in a consonant which is considered to be an extrametrical syllable, denoted by parenthesised $\sigma$, i.e. ($\sigma$). Based on this notion, the nominal stems in (30) are analysed moraically in (34), showing syllable-weight patterns.

(34) MORAIC ANALYSIS OF NOMINAL STEMS

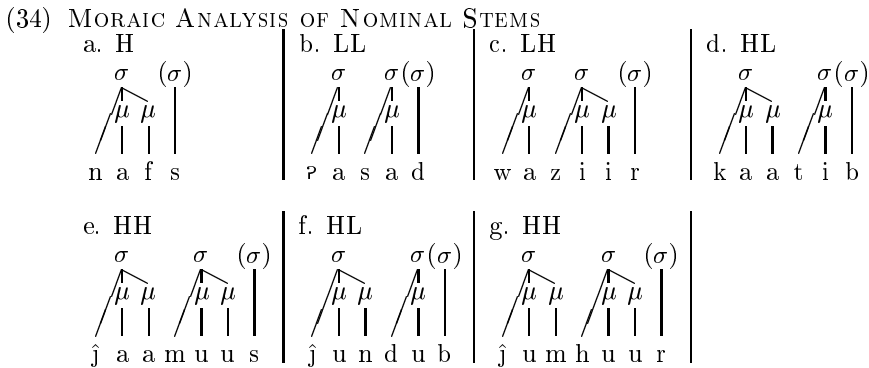

The classification indicates that the Arabic nominal stem consists of at least two morae and at most two syllables, in addition to the obligatory final ($\sigma$). This statement is given formally in (35).

(35) MINIMAL AND MAXIMAL STEM CONSTRAINT
    a. Nominal templates are minimally bimoraic.
    b. Nominal templates are maximally bisyllabic.

One aspect of this analysis has not been mentioned yet: template satisfaction; i.e. the association of melodic elements with templates, especially since moraic theory does not distinguish between CVV and CVC syllables. The final syllable of a stem is predictable: a bimoraic final syllable is CVC in monosyllabic stems, e.g. (34a), and CVV in disyllabic ones, e.g. (34c,e,g). The initial syllable is not predictable, but can be determined from the length of the root morpheme: a bimoraic initial syllable is CVV in trilateral roots, e.g. (34e), and CVC in quadrilateral roots, e.g. (34f-g).

McCarthy (1993) claims that there are no HL noun templates in Arabic; in other words, the representations in (34d,f) are a-templatic. The majority of CVVCVC trilateral forms like /kaatib/ are active participles of the verb (Measure 1), e.g. /kaatib/ is derived from the verb /katab/ by affixing a mora $\mu$ to the first syllable, and providing a new vowel melody (this is similar to the derivation of /kaatab/ (Measure 3) in section 5.2.2). Forms like /xaatam/ 'signet-ring' are rare. Quadrilateral forms like /jundub/ (34f) are a-templatic; they are formed by organising the Cs and Vs in accordance to well-formedness in Arabic syllable structure.

---

[7]Other conventions associate consonant melodies left-to-right to the moraic nodes, followed by associating vowel melodies to syllable-initial morae.

To summarise: Templates are described by the units of prosody, i.e. syllable and mora. Syllables in Arabic are three kinds: light CV (monomoraic, $\sigma_\mu$), heavy CVC (bimoraic, $\sigma_{\mu\mu}$) and heavy CVV (bimoraic, $\sigma_{\mu\mu}$); no distinction is made between CVV and CVC syllables. All stems in Arabic end in an obligatory consonant. Nominal stems are templatic, but there is no HL template; forms like /kaatib/ are active participle of the verb (Measure 1) and quadrilateral forms like /ĵundub/ are sequences of consonants and vowels.

## 4.2 Computational Analysis

This section provides a computational analysis of the moraic account. The two-level grammar makes use of the fact that the chosen formalism allows sequences of unequal lengths. (The analysis of a-templatic forms like /kaatib/ are similar to the derivation of the verb /kaatab/ (Measure 3), shown in section 5.2.2.)

### 4.2.1 The Lexica

The moraic lexicon makes use of three tapes: pattern, root and vocalism. However, in order to use the same lexicon when discussing the broken plural (section 6), four tapes will be used. The fourth tape will be empty at this stage. Sample lexica appear in (36).

(36) MORAIC LEXICA
   Number of Trees: 4.
   a. Pattern Lexicon (on Tree 1):
      1  $\{\sigma_{\mu\mu}\sigma\}$        pattern:[measure=h, number=N]
      2  $\{\sigma_\mu\sigma_\mu\sigma\}$      pattern:[measure=ll, number=N]
      3  $\{\sigma_\mu\sigma_{\mu\mu}\sigma\}$     pattern:[measure=lh, number=N]
      4  $\{\sigma_{\mu\mu}\sigma_{\mu\mu}\sigma\}$    pattern:[measure=hh, number=N]
   b. Root Lexicon (on Tree 2):
      1  {nfs} 'soul'          root:[measure=h, sing_vowel=a, pl_vowel=u]
      2  {rĵl} 'man'           root:[measure=ll, sing_vowel=au, pl_vowel=ia]
      3  {wzr} 'minister'      root:[measure=lh, sing_vowel=ai]
      4  {ĵms} 'buffalo'       root:[measure=hh, sing_vowel=au]
      5  {ĵmhr} 'multitude'    root:[measure=hh, sing_vowel=u]
   c. Vocalism Lexicon (on Tree 3):
      1  {a}    vocalism:[sing_vowel=a]
      2  {ai}   vocalism:[sing_vowel=ai]
      3  {au}   vocalism:[sing_vowel=au]
   d. A-templatic Lexicon (on Tree 1):
      1  {ĵundub} 'locust'        noun_stem:[number=N]
      2  {xaatam} 'signet-ring'   noun_stem:[number=N]

The pattern lexicon provides the four templates H, LL, LH and HH; note that $\sigma_\mu$ and $\sigma_{\mu\mu}$ are atomic elements, and extrametrical syllables are denoted by $\sigma$. Each pattern morpheme is associated with a feature structure which indicates the morpheme's measure and number. The value of the latter is a variable N which unifies with rules associated with the feature structures [number=sing] or [number=pl]; this is similar to the German 'umlaut' problem (Trost 1990).

The root lexicon provides the roots of the words mentioned in the above discussion; the feature structure indicates the root's measure and singular vocalism (bimoraic stems also provide the plural vocalism which will be used in section 6). The vocalism lexicon provides the various singular vocalisms (other vowel melodies are not shown). Finally, the a-templatic lexicon provides a-templatic stems.

### 4.2.2 The Grammar

Each lexical expression in the moraic grammar is a 4-tuple (the 4th item is used for the broken plural in section 6.2.2). A two-level grammar which describes the stems cited above appears in (37).

(37) MORAIC GRAMMAR
    a. General Rules:
        R1   Identity:              * - X - *   $\Rightarrow$   * - X - *
        R2   Extrametricality:    * - C - *   $\Leftrightarrow$   * - ($\sigma$,C, , ) - (+,+,+, )
        R3   Monomoraic CV:   * - CV - *   $\Rightarrow$   * - ($\sigma_\mu$,C,V), - *   [number=sing]
    b. Initial Bimoraic Rules:
        R4   CVC:   * - $C_1VC_2$ - *   $\Rightarrow$   * - ($\sigma_{\mu\mu}$,$C_1C_2$,V, ) - ($\sigma_{\mu\mu}$,*,*, )   [number=sing]
        R5   CVV:   * - CVV - *   $\Rightarrow$   * - ($\sigma_{\mu\mu}$,C,V, ) - ($\sigma_{\mu\mu}$,*,*, )   [number=sing]
    c. Final Bimoraic Rules:
        R6   CVC:   * - $C_1VC_2$ - *   $\Leftrightarrow$   * - ($\sigma_{\mu\mu}$,$C_1C_2$,V, ) - ($\sigma$,*, , )   [number=sing]
        R7   CVV:   * - CVV - *   $\Leftrightarrow$   (S,*,*, ) - ($\sigma_{\mu\mu}$,C,V, ) - ($\sigma$,*, , )   [number=sing]
    d. Boundary Rules:
        R8   Non-stem:   * -  - *   $\Rightarrow$   (A, , , ) - + - *
        R9   Stem:   * -  - *   $\Rightarrow$   (A,*, , ) - (+,+,+, ) - *
    e. Spreading Rules:
        R10   Monomoraic spreading:   * - CV - *   $\Rightarrow$   (S,*,V, ) - ($\sigma_\mu$,C, , ) - *
        R11   Bimoraic spreading:   * - CVV - *   $\Rightarrow$   (S,*,V, ) - ($\sigma_{\mu\mu}$,C, , ) - *
    where: $C_1,C_2$=`radical`; V=`vowel`; S $\in \{\sigma_\mu,\sigma_{\mu\mu}\}$; A $\neq$ +.

R1 is the identity rule which allows any lexical character on the first tape to surface. R2 is an obligatory rule which states that reading $\sigma$ from the first tape and a C from the second tape maps to C on the surface; this accounts for the obligatory final ($\sigma$) which must be followed by lexical boundary symbols, i.e. RLC = (+,+,+, ). R3 is the monomoraic rule which reads $\sigma_\mu$ from the first tape, a C from the second tape and a V from the third tape, mapping them to CV on the surface.

Rules R4-R5 sanction initial bimoraic syllables. R4 handles CVC syllables by reading a $\sigma_{\mu\mu}$ from the first tape, two characters $C_1C_2$ from the second tape, and a V from the third tape, mapping them to $C_1VC_2$ on the surface. R5 handles CVV syllables reading $\sigma_{\mu\mu}$ from the first tape, C from the second and V from the third, mapping them to CVV on the surface. In both rules, RLC = ($\sigma_{\mu\mu}$,*,*, ) ensures that they only apply to the initial syllable of a stem. There is no need to code the fact that a bimoraic initial syllable is CVC in trilateral roots and CVV in quadrilateral ones; if a rule applies on the wrong root, the analysis will be doomed to fail since there will be one extra (or one less) character in the lexical entry.

Rules R6-R7 sanction final bimoraic syllables. Their LEX and SURF are equivalent to R4 and R5, respectively, but they differ in the context LLC = ($\sigma$,*,*, ) which ensures that they are applied to the final syllable of a stem. Recall that final bimoraic syllables are realised as CVC in monosyllabic stems and CVV in bisyllabic ones; hence, the difference between R6 and R7 lies in RLC and SURF: R6 applies to monosyllabic stems and R7 applies to bisyllabic stems by virtue of their RLC. Both rules are obligatory.

By setting the lexical contexts in R4-R7 as shown above, the rules ensure that (i) HL templates are not allowed, and (ii) the minimal and maximal constraints are respected. Further, R3-R7 are associated with the feature structure `[number=sing]` which must unify with lexical feature on the first tape.

The boundary rules in (37d) are the same as those in the CV grammar. The spreading rules in (37e) allow intersyllabic spreading. Note that intrasyllabic spreading is dealt with in R5 and R7 by vowel lengthening. The two-level analysis of the cited data is illustrated in (38).

(38) MULTI-TAPE TWO-LEVEL MORAIC ANALYSIS

a. H

| a |   | + | VT |
|---|---|---|----|
| nf | s | + | RT |
| $\sigma_{\mu\mu}$ | $\sigma$ | + | PT |
| 6 | 2 | 9 | |
| naf | s | | ST |

b. LL

| a |   |   | + | VT |
|---|---|---|---|----|
| ʔ | s | d | + | RT |
| $\sigma_\mu$ | $\sigma_\mu$ | $\sigma$ | + | PT |
| 3 | 10 | 2 | 9 | |
| ʔa | sa | d | | ST |

c. LH

| a | i |   | + | VT |
|---|---|---|---|----|
| w | z | r | + | RT |
| $\sigma_\mu$ | $\sigma_{\mu\mu}$ | $\sigma$ | + | PT |
| 3 | 7 | 2 | 9 | |
| wa | zii | r | | ST |

d. HH

| a | u |   | + | VT |
|---|---|---|---|----|
| ĵ | m | s | + | RT |
| $\sigma_{\mu\mu}$ | $\sigma_{\mu\mu}$ | $\sigma$ | + | PT |
| 5 | 7 | 2 | 9 | |
| ĵaa | muu | s | | ST |

e. HH

| u |   |   | + | VT |
|---|---|---|---|----|
| ĵm | h | r | + | RT |
| $\sigma_{\mu\mu}$ | $\sigma_{\mu\mu}$ | $\sigma$ | + | PT |
| 4 | 11 | 2 | 9 | |
| ĵum | huu | r | | ST |

f. a-templatic stem

| ĵ | u | n | d | u | b | + | Lex |
|---|---|---|---|---|---|---|-----|
| 1 | 1 | 1 | 1 | 1 | 8 | | |
| ĵ | u | n | d | u | b | | ST |

The numbers between the two levels indicate the rule numbers in (37) which sanction the sequences. The morphosyntactic grammar appears in (39).

(39) MORAIC MORPHOTACTIC GRAMMAR
```
noun_stem:[measure=M, number=N] ⟶
    pattern:[measure=M, number=N]
    root:[measure=M, sing_vowel=SV]
    vocalism:[sing_vowel=SV].
```

The value of N is realised from rule feature structures.

To summarise: The pattern lexicon consists of moraic templates. The remaining lexica (i.e. root and vocalism) are similar to those in the CV grammar. A-templatic stems are entered in a separate lexicon which sits on the first lexical tape.

The grammar sanctions one syllable at a time in the stem. Lexical contexts ensure that HL templates are not allowed; they also encode the minimal and maximal constrains. Since nominals are maximally bisyllabic, spreading rules do not require ellipses because intersyllabic spreading occurs between adjacent syllables. Intrasyllabic spreading is dealt with by vowel lengthening.

## 5  AFFIXATIONAL ANALYSIS

The affixational analysis presented here is applied to the Arabic verb. Section 5.1 describes McCarthy's new proposal, followed by a computational account in section 5.2.

### 5.1  Linguistic Description

In a recent study, McCarthy (1993) departed radically from the notion of root-and-pattern morphology in the Arabic verbal stem. The new proposal follows: Arabic Measure 1, /katab/, is templatic having the template in (40).   M 1

(40) ARABIC MEASURE 1 TEMPLATE

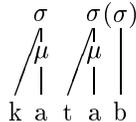

The remaining measures are derived from the Measure 1 template by affixation; they have no templates of their own. The simplest operation is prefixation, e.g. {n} + Measure 1 → /nkatab/   M 7
(Measure 7). Measures 4 and 10 are derived in a similar fashion, but undergo a rule of syncope as   M 4, 10
shown in (41).

(41) DERIVATION OF MEASURES 4 AND 10
Syncope: V ⟶ φ /[CVC ___ CVC]$_{stem}$
a. Measure 4: ʔa + katab ⟶ */ʔakatab/ $\stackrel{syncope}{\longrightarrow}$ /ʔaktab/
b. Measure 10: sta + katab ⟶ */stakatab/ $\stackrel{syncope}{\longrightarrow}$ /staktab/

The remaining measures are derived by affixation under prosodic circumscription, a notion introduced by McCarthy and Prince (1990a). Prosodic circumscription allows a morphological rule to apply on a prosodically-delimited substring within a stem. Let B be a base stem (string) on which a morphological operation takes place; the parsing function Φ <C, E> returns the substring denoted by the constituent C from the edge E of the base B, where E ∈ {right, left}. For example, the function Φ <CVC, right> returns the closed syllable at the right edge of B; this substring is denoted by B:Φ. The remaining substring of B, i.e. the residue, is denoted by B/Φ. To illustrate this, let B = /katab/; applying the function Φ<CVC, right> on /katab/ splits it into two substrings: (i) B:Φ = /tab/ is the closed syllable at the right edge of B, and (ii) B/Φ = /ka/ is the residue of B.

A morphological operation is denoted by O, e.g. O = 'prefix {t}'. There are two types of prosodic circumscription: positive and negative. In positive prosodic circumscription, the operation O applies to B:Φ; this type is denoted by O:Φ <C,E>. In negative prosodic circumscription, the operation applies to the residue B/Φ; this type is denoted by O/Φ <C,E>.

Negative prosodic circumscription is illustrated by the derivation of /ktatab/ (Measure 8) by the affixation of a {t} to the base template /katab/. The operation is O = 'prefix {t}'. The rule is O/Φ <C, left> which states that a morpheme {t} is to be prefixed to the residue of the base; the residue is obtained after stripping the first consonant C from the left edge. The process is illustrated in (42).    M 8

(42) DERIVATION OF MEASURE 8
| Base | katab |
| Negative Circumscription | <k> atab |
| Prefix {t} | <k> t + atab |
| Result | ktatab |

Measure 2 is derived by prefixing a mora under negative prosodic circumscription. The operation is O = 'prefix μ' and the rule is O/Φ <μ, left>. The new mora is filled by the spreading of the following consonant. The derivation is illustrated in (43).    M 2

(43) DERIVATION OF MEASURE 2
| Base | katab |
| Negative Circumscription | <ka> tab |
| Prefix μ | <ka> μ + tab |
| Spread | <ka> t tab |
| Result | kattab |

Measure 3 is derived by prefixing the LL base template with μ. The process is illustrated in (44).    M 3

(44) DERIVATION OF /kaatib/

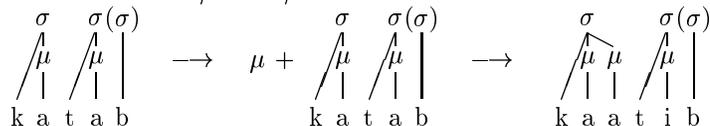

The remaining (rare) measures occur in Wehr's dictionary between 0-8 times, apart from Measure 9 which occurs 18 times. Most of them can be analysed in similar lines by pre-associating the second radical with the first σ node of the LL template.

To summarise: Measure 1 constitutes the base LL template from which all other measures are derived by affixation. (45) tabulates the derivations of common measures.

(45) SUMMARY OF VERBAL DERIVATIONS

| Measure | Verb | Derivation |
|---|---|---|
| 1 | katab | = LL base template |
| 2 | kattab | $\mu$ + Measure 1 under O/$\Phi$ <$\mu$, left> |
| 3 | kaatab | $\mu$ + Measure 1 |
| 4 | ʔaktab | {ʔa} + Measure 1 (syncope applies) |
| 5 | takattab | {ta} + Measure 2 |
| 6 | takaatab | {ta} + Measure 3 |
| 7 | nkatab | {n} + Measure 1 |
| 8 | ktatab | {t} + Measure 1 under O/$\Phi$ <C, left> |
| 10 | staktab | {sta} + Measure 1 (syncope applies) |

Rare measures are probably a peripheral phenomenon, but can, in most cases, be catered for with prosodic circumscription.

## 5.2 Computational Analysis

This section provides a computational account of the affixational model describing the Arabic perfect verb. The two-level grammar presented in section 5.2.2 does not model prosodic circumscription *per se*. What it aims to do is to demonstrate how prosodic circumscription can be catered for under two-level theory.

### 5.2.1 The Lexica

The affixational lexicon maintains four tapes: pattern, root, vocalism and affix tapes. Sample lexica appear in (46).

(46) AFFIXATIONAL LEXICA
   No of Trees: 4.
   a. Pattern Lexicon (on Tree 1):
      1   {$\sigma_\mu \sigma_\mu \sigma$}   pattern:[measure=(1-8,10)]
   b. Verbal Affixes Lexicon (on Tree 4):
      1   {ʔV}    verb_affix:[measure=4]
      2   {n}     verb_affix:[measure=7]
      3   {tV}    verb_affix:[measure=(5,6)]
      4   {t}     verb_affix:[measure=8]
      5   {stV}   verb_affix:[measure=10]
   c. Root Lexicon (on Tree 2):
      1   {ktb} 'to write'      root:[measure=(1,2,3,4,6,7,8,10), perf_vowel=a]
      2   {smħ} 'be generous'   root:[measure=(1,2,3,6,10), perf_vowel=u]
      3   {smʕ} 'to hear'       root:[measure=(1,2,4,5,6,8), perf_vowel=i]
   d. Vocalism Lexicon (on Tree 3):
      1   {a}    vocalism:[tense=perf, voice=act, perf_vowel=a]
      2   {au}   vocalism:[measure=1, tense=perf, voice=act, perf_vowel=u]
      3   {ai}   vocalism:[measure=1, tense=perf, voice=act, perf_vowel=i]
      4   {ui}   vocalism:[tense=perf, voice=pass]
   e. Rare Verbs Lexicon (on Tree 1):
      1   {swadad} 'to be black'    verb_stem:[measure=9, tense=perf, voice=act]
      2   {ħmaarar} 'to be red'     verb_stem:[measure=11, tense=perf, voice=act]
      3   {ħdawdab} 'to be arched'  verb_stem:[measure=12, tense=perf, voice=act]
      4   {ʕlawwad} 'to be heavy'   verb_stem:[measure=13, tense=perf, voice=act]
      5   {sħankak} 'to be dark'    verb_stem:[measure=14, tense=perf, voice=act]
      6   {ʕlanday} 'to be strong'  verb_stem:[measure=15, tense=perf, voice=act]

(46a) provides the pattern lexicon which maintains the one LL base template. (46b) provides the verbal affix morphemes. Notice that the vowel in the affixes of Measures 4, 5-6 and 10 is a variable

V. This makes it possible for the affix to have a different vowel according to the mood of the following stem, e.g. [a] in /takattab/ and [u] in /tukuttib/ (Measure 5). The root and vocalism lexica are similar to the ones appearing in the CV lexicon (section 3.2.1). The rare lexicon lists rare verbs with their measure numbers, from Wright (1988).[8]

### 5.2.2 The Grammar

Each lexical expression in the affixational two-level grammar is a 4-tuple. A sample grammar appears in (47).

(47) AFFIXATIONAL GRAMMAR
    a. General Rules:
        R1    Identity:                  * - X - *   $\Rightarrow$   * - X - *
        R2    Extrametricality:       * - C - *   $\Rightarrow$   * - ($\sigma$,C, , ) - (+,+,+, )
        R3    Monomoraic syllable:   * - CV - *   $\Rightarrow$   * - ($\sigma_\mu$,C,V, ) - *
        R4    Affixes:                   * - A - *   $\Rightarrow$   * - ( , , ,A) - *
    b. Boundary Rules:
        R5    Non-stem:    * -   - *   $\Rightarrow$   (B, , ) - + - *
        R6    Stem:             * -   - *   $\Rightarrow$   (B,*, , ) - (+,+,+, ) - *
        R7    Affix:             * -   - *   $\Rightarrow$   (*,*,*,B) - ( , , ,+) - *
    c. Syncope Rule:
        R8    Syncope:   ($C_1$,V) - C - ($C_2$,$V_1$,$C_3$)   $\Leftrightarrow$   * - ($\sigma_\mu$,C,V, ) - *
    d. Measure-Specific Rules:
        R9     Measures 2/5:   * - C - *   $\Rightarrow$   ($\sigma_\mu$,$C_1$,$V_1$, ) -   - ($\sigma_\mu$,C,*, )  [measure=(2,5)]
        R10   Measures 3/6:   * - CVV - *   $\Rightarrow$   * - ($\sigma_\mu$,C,V, ) - *  [measure=(3,6)]
        R11   Measure 8:     * - CAV - *   $\Rightarrow$   * - ($\sigma_\mu$,C,V,A) - *  [measure=8]
    e. Spreading Rules:
        R12   Spreading:   * - CV - *   $\Rightarrow$   ($\sigma_\mu$,*,V,*) $\cdots$ - ($\sigma_\mu$,C, , ) - ($\sigma$,*, , )
    where: $C_i$=`radical`; $V_i$=`vowel`; A=`verbal affix`; B $\neq$ +.

Rules R1-R3 are similar to the ones appearing in the moraic grammar (section 4.2.2); so are R5-R6 (R8-R9 in the moraic grammar). R4 and R7 take care of affixes: the former allows an affix character which appears on the fourth lexical tape to surface, and the latter caters for the affix morpheme boundary symbol.

R8 is the syncope rule; note that V in LSC must unify with V in LEX ensuring that the vowel of the affix has the same quality as that of the stem, e.g. /ta+kattab/ and /tu+kuttib/ (Measure 5).

The rules in (47d) are measure-specific. Each rule is associated with a feature structure which must unify with the feature structures of the affected lexical entries. This makes sure that each rule is applied only to the proper measure. R9 handles /kattab/ (Measure 2) and /takattab/ (Measure 5). It represents the operation O = 'prefix $\mu$' and the rule O/$\Phi$ <$\mu$, `left`> by placing B:$\Phi$ in LLC and the residue B/$\Phi$ in RLC, and inserting a consonant C (representing $\mu$) on the surface. The filling of $\mu$ by the spreading of the second radical is achieved by the unification of C in SURF with C in RLC. R10 handles /kaatab/ (Measure 3) and /takaatab/ (Measure 6). It adds a $\mu$ by lengthening the vowel V into VV. R11 takes care of /ktatab/ (Measure 8). It represents the operation O = 'prefix {t}' and the rule O/$\Phi$ <C, `left`>.

Finally, the spreading rule allows the active vowel [a] to spread over the stem. The derivations of the cited measures appear in (48).

---

[8] All rare forms surface with prosthetic ʔV. In measures 9 and 11, final CVC] $\rightarrow$ CC. The cited verbs are derived from the roots {swd}, {ḥmr}, {ḥdb}, {ʕld}, {sḥk} and {ʕld}, respectively.

(48) MULTI-TAPE DERIVATIONS OF THE ARABIC VERB

Measure 1

| a |   |   | + |   | $VT$ |
|---|---|---|---|---|------|
| k | t | b | + |   | $RT$ |
| $\sigma_\mu$ | $\sigma_\mu$ | $\sigma$ | + |   | $PT$ |
| 3 | 12 | 2 | 6 |   |      |
| ka | ta | b |   |   | $ST$ |

Measure 2

| a |   |   |   | + | $VT$ |
|---|---|---|---|---|------|
| k |   | t | b | + | $RT$ |
| $\sigma_\mu$ |   | $\sigma_\mu$ | $\sigma$ | + | $PT$ |
| 3 | 9 | 12 | 2 | 6 |    |
| ka | t | ta | b |   | $ST$ |

Measure 3

| a |   |   | + |   | $VT$ |
|---|---|---|---|---|------|
| k | t | b | + |   | $RT$ |
| $\sigma_\mu$ | $\sigma_\mu$ | $\sigma$ | + |   | $PT$ |
| 10 | 12 | 2 | 6 |   |      |
| kaa | ta | b |   |   | $ST$ |

Measure 4

| ʔ | a | + |   |   |   |   | $AT$ |
|---|---|---|---|---|---|---|------|
|   |   | a |   |   |   | + | $VT$ |
|   |   | k | t | b | + |   | $RT$ |
|   |   | $\sigma_\mu$ | $\sigma_\mu$ | $\sigma$ | + |   | $PT$ |
| 4 | 4 | 7 | 8 | 12 | 2 | 6 |    |
| ʔ | a |   | k | ta | b |   | $ST$ |

Measure 5

| t | a | + |   |   |   |   |   | $AT$ |
|---|---|---|---|---|---|---|---|------|
|   |   | a |   |   |   |   | + | $VT$ |
|   |   | k |   | t | b | + |   | $RT$ |
|   |   | $\sigma_\mu$ |   | $\sigma_\mu$ | $\sigma$ | + |   | $PT$ |
| 4 | 4 | 7 | 3 | 9 | 12 | 2 | 6 |    |
| t | a |   | ka | t | ta | b |   | $ST$ |

Measure 8

| t | + |   |   |   |   | $AT$ |
|---|---|---|---|---|---|------|
| a |   |   |   |   | + | $VT$ |
| k |   | t | b | + |   | $RT$ |
| $\sigma_\mu$ |   | $\sigma_\mu$ | $\sigma$ | + |   | $PT$ |
| 11 | 7 | 12 | 2 | 6 |   |    |
| kta |   | ta | b |   |   | $ST$ |

Measure 7

| n | + |   |   |   |   | $AT$ |
|---|---|---|---|---|---|------|
|   | a |   |   |   | + | $VT$ |
|   | k | t | b | + |   | $RT$ |
|   | $\sigma_\mu$ | $\sigma_\mu$ | $\sigma$ | + |   | $PT$ |
| 4 | 7 | 3 | 12 | 2 | 6 |    |
| n |   | ka | ta | b |   | $ST$ |

Measure 10

| s | t | a | + |   |   |   |   | $AT$ |
|---|---|---|---|---|---|---|---|------|
|   |   |   | a |   |   |   | + | $VT$ |
|   |   |   | k | t | b | + |   | $RT$ |
|   |   |   | $\sigma_\mu$ | $\sigma_\mu$ | $\sigma$ | + |   | $PT$ |
| 4 | 4 | 4 | 7 | 8 | 12 | 2 | 6 |    |
| s | t | a |   | k | ta | b |   | $ST$ |

The morphosyntactic grammar is shown in (49). The rule 'verb → verb_stem vim' is the same as the one in section 4.2.2.

(49) AFFIXATIONAL MORPHOSYNTACTIC GRAMMAR
   verb_stem:[measure=M, tense=T, voice=V] ⟶
      pattern:[measure=M, tense=T]
      root:[measure=M, perf_vowel=PV]
      vocalism:[tense=T, voice=V, perf_vowel=PV].
   verb_stem:[measure=M, tense=T, voice=V] ⟶
      verb_affix:[measure=M]
      verb_stem:[measure=M, tense=T, voice=V].

To summarise: The pattern lexicon consists of the one LL base template. Affixes are entered in a unique lexicon which sits on its own lexical tape. The root and vocalism lexica are similar to those in earlier grammars. A-templatic stems are entered in a separate lexicon and are placed on the first lexical tape. The syllabic rules of the affixational grammar are similar to those of the moraic grammar. Measures derived by prosodic circumscription have their own unique rules.

# 6   THE BROKEN PLURAL

Finally, we discuss the challenging phenomenon of the broken plural.[9] Various nominal classes appear in (50) (McCarthy and Prince 1990a).

---

[9]I know of no previous computational treatment of the Arabic broken plural, apart from a 10-state FSM mentioned in Sproat (1992: p. 160).

(50) ARABIC BROKEN PLURAL FORMS

|   | Singular | Plural |   | Singular | Plural |
|---|---|---|---|---|---|
| a. | CVCC | | d. | CVVCVC+at | |
| | nafs 'soul' | nufuus | | faakih 'fruit' | fawaakih |
| | qidħ 'arrow' | qidaaħ | | ʔaanis 'cheerful' | ʔawaanis |
| | ħukm 'judgement' | ħakaam | | | |
| b. | CVCVC | | e. | CVVCV(V)C | |
| | ʔasad 'lion' | ʔusuud | | xaatam 'signet-ring' | xawaatim |
| | raĵul 'man' | riĵaal | | ĵaamuus 'buffalo' | ĵawaamiis |
| | ʕinab 'grape' | ʕanaab | | | |
| c. | CVCVVC+at | | f. | CVCCV(V)C | |
| | saħaab 'cloud' | saħaaʔib | | ĵundub 'locust' | ĵanaadib |
| | ĵaziir 'island' | ĵazaaʔir | | sultˤaan 'sultan' | salaatˤiin |
| | ħaluub 'milch-camel' | ħalaaʔib | | | |

/ħakaam/ (50a) and /ʕanaab/ (50b) surface as /ʔaħkaam/ and /ʔaʕnaab/, respectively, by Ca Metathesis.

Section 6.1 describes the linguistic analysis of the broken plural, followed by a computational account in section 6.2.

## 6.1 Linguistic Description

McCarthy and Prince (1990a) argued that the broken plural in Arabic is derived from the singular stem, not from the root; hence, /ĵundub/ 'singular' → /ĵanaadib/ 'plural'. They show that broken plurals have the iambic template given in (51).

(51) IAMBIC TEMPLATE

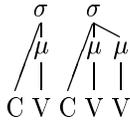

An informal account of the derivation of the plural is illustrated by deriving /ĵanaadib/ from /ĵundub/. Both forms are given in (52).

(52) /ĵundub/ AND /ĵanaadib/
  a. Singular   b. Plural

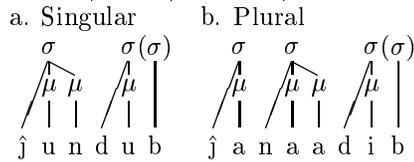

The derivation takes the following line: the first two morae of the singular, i.e. /ĵun/, are mapped onto the iambic template in (51); the consonants link to syllable nodes and the vowel [u] spreads over the three morae yielding the representation in (53a). Now the plural melody {ai} overwrites the singular melody by the spreading of [a] yielding the representation in (53b); notice that the final vowel [i] remains unlinked. Finally, the remaining part of the singular, i.e. /dub/, is added and the unlinked [i] overwrites the original vowel [u]. The final representation is shown in (53c).

(53) DERIVATION OF /ĵanaadib/
  a.   b.   c.

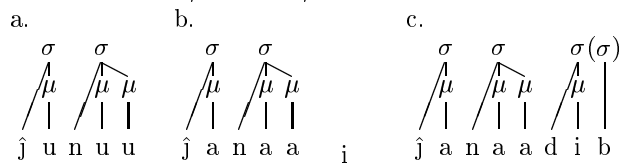

This process is accomplished by positive prosodic circumscription, discussed earlier in section 5.1. The constituent C in the parsing function Φ <C, Edge> is the minimal word, denoted by $W_{min}$, which consists of two morae; since melody mapping in Arabic is left-to-right, Edge = left. The parsing function is then Φ <$W_{min}$, left>. After the function parses a singular stem B, it returns B:Φ which maps onto the iambic template. The residue B/Φ is then added and the plural melody overwrites the singular one. The parsing results of the examples cited above are listed in (54).

(54) DERIVATION OF THE BROKEN PLURALS

| | Singular | B:Φ | B/Φ | Plural |
|---|---|---|---|---|
| a. | CVCC | | | |
| | nafs | naf | s | nufuu+s |
| | qidħ | qid | ħ | qidaa+ħ |
| | ħukm | ħuk | m | ħakaa+m (surfaces as /ʔaħkaam/) |
| b. | CVCVC | | | |
| | ʔasad | ʔasa | d | ʔusuu+d |
| | raĵul | raĵu | l | riĵaa+l |
| | ʕinab | ʕina | b | ʕanaa+b (surfaces as /ʔaʕnaab/) |
| c. | CVCVVC+at | | | |
| | saħaab | saħa | ab | saħaa+ʔib |
| | ĵaziir | ĵazi | ir | ĵazaa+ʔir |
| | ħaluub | ħalu | ub | ħalaa+ʔib |
| d. | CVVCVC+at | | | |
| | faakih | faa | kih | fawaakih |
| | ʔaanis | ʔaa | nis | ʔawaanis |
| e. | CVVCV(V)C | | | |
| | xaatam | xaa | tam | xawaa+tim |
| | ĵaamuus | ĵaa | muus | ĵawaa+miis |
| f. | CVCCV(V)C | | | |
| | ĵundub | ĵun | dub | ĵanaa+dib |
| | sulṭaan | sul | ṭaan | salaa+ṭiin |

Starting from the bottom of the list, the derivations of the forms in (54f) follow the example in (53). The derivation of the forms in (54e), e.g. /ĵaamuus/ → /ĵawaamiis/, is illustrated in (55).

(55) DERIVATION OF /ĵawaamiis/

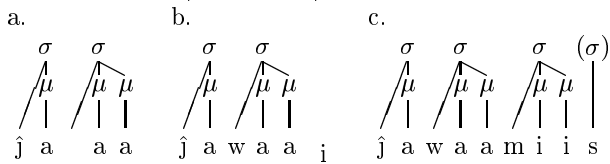

B:Φ, i.e. /ĵaa/, maps to the iambic template as illustrated in (55a), with the spreading of [a]. Since there is no consonant to fill the second σ node, a [w] is inserted as shown in (55b); the insertion of [w] can be expressed by the rule in (56).

(56) CONSONANTAL DEFAULT RULE
    $\phi \to w$, when required by syllabic well-formedness.

Finally, B/Φ, i.e. /miis/, is added as shown earlier; the result appears in (55c). Notice that the length of [i] is carried over from the singular stem. The forms in (54d) are derived in a similar manner.

The forms in (54c) add some complications. Here, the parsing function splits the second syllable into two parts, e.g. Φ(saħaab) results in B:Φ = /saħa/ and B/Φ = /ab/. The former maps to the iambic template as in earlier examples; this is illustrated in (57a).

(57) DERIVATION OF /saħaaʔib/

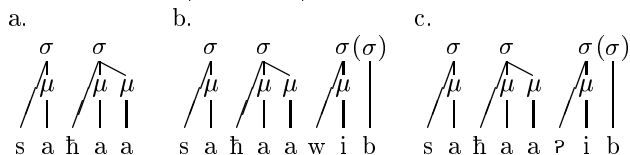

a.    b.    c.

Next, B/Φ is added, the vowel melody is overwritten by the plural one, and rule (56) is invoked since B/Φ does not constitute a syllable, resulting in the σ node being linked to [w] as shown in (57b). Finally, a phonological rule of glide realization, where [w] → [ʔ], is applied to derive the surface form.

The derivation of plurals from bimoraic stems (54a-b) is straightforward; however, (i) the plural vowel melody may be one of four: {u}, {ia}, {a} or {au} and is lexically marked; (ii) since B/Φ is an extraprosodic syllable, the [i] in the plural vowel melody {ai} is deleted by Stray Erasure. The derivation of /ħakaam/ is illustrated in (58a-b).

(58) DERIVATION OF /ħakaam/

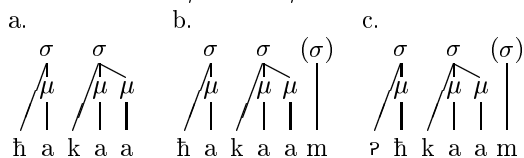

a.    b.    c.

Forms like /ħakaam/ and /ʕanaab/ surface as /ʔaħkaam/ and /ʔaʕnaab/ by a Ca metathesis rule; the result appears in (58c).

To summarise: broken plurals are derived from their respective singular stems - not roots - by positive prosodic circumscription. The parsing function Φ <W$_{min}$, `left`> splits a singular stem B into B:Φ and the residue B/Φ. The process continues according to the procedure in (59).

(59) PROCEDURE FOR DERIVING THE BROKEN PLURAL
   a. Map B:Φ to the iambic template.
   b. Concatenate B/Φ to the result.
   c. Replace the vowel melody with the plural melody {ai} (or with the lexically marked melody in the case of monosyllabic singular).
   d. If a σ node is not linked, associate it with [w]; under some phonological environments, [w] → [ʔ].
   e. In certain phonological environments, apply Ca Metathesis.

## 6.2 Computational Analysis

Because of its dependency on the singular (which does not constitute a lexical entry), the broken plural seems to pose a theoretical 'hazard' for two-level theory. Consider the following facts: firstly, the length of the final syllable vowel is carried from the singular to the plural, e.g. /ĵund**u**b/ → /ĵanaad**i**b/, /ṣulṭ**aa**n/ → /ṣalaaṭ**ii**n/; secondly, the number of syllables in the plural depends on the number of morae in the singular (bimoraic stems form disyllabic plurals, and longer stems form trisyllabic plurals); thirdly, tri-consonantal singular with a long vowel require the insertion of [w], realised as [ʔ] under certain phonological conditions (McCarthy and Prince 1990a: p. 218). In other words, the derivation of the broken plural takes the following form: root → singular → plural. Such a derivation requires a *three*-level model as illustrated in (60a). The broken plural is derived here from the root via the **implicit derivation** of the singular using our two-level model. The concept is illustrated in (60b).

(60) TWO-LEVEL VS THREE-LEVEL DERIVATION
   a. Three level System   b. Two-level System

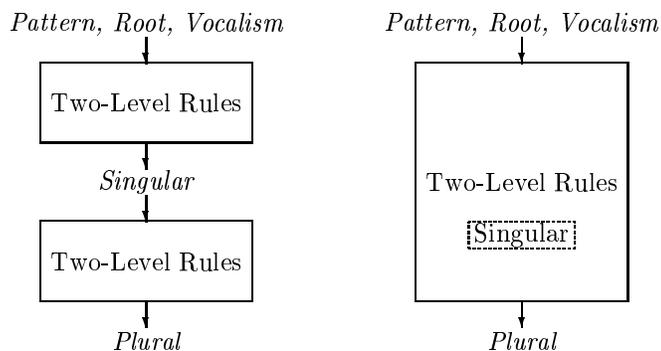

Deriving the broken plural via the implicit derivation of the singular is best explained by an example. To derive /ĵawaamiis/ (singular /ĵaamuus/), a two-level rule derives B:Φ, i.e. /ĵaa/, from lexical morphemes along the lines of the moraic grammar in section 4.2.2. However, instead of mapping B:Φ to /ĵaa/ on the surface, the rule maps B:Φ to the iambic plural template CVCVV, i.e. /ĵawaa/ (the rule also takes care of replacing the singular vowel melody with the plural one, and filling the second C slot with the default [w], if needed). A second rule maps the rest of the morpheme characters, i.e. the residue B/Φ, to the surface as they are, but overwriting the singular vowel melody with the plural one. The derivation, then, respects two-level theory. We say that the plural is derived via the 'implicit derivation' of the singular because the two-level rules find a singular form, but map it to the corresponding plural form on the surface. More detailed examples appear in section 6.2.2.

### 6.2.1 The Lexicon

The broken plural lexicon builds on the moraic lexicon of section 4.2.1. The additional morphemes are given in (61).

(61) BROKEN PLURAL LEXICON
   Plural Vocalism Lexicon (on Tree 4):
   1   {aa}   pl_vocalism:[measure=(h,ll), pl_vowel=a]
   2   {ia}   pl_vocalism:[measure=(h,ll), pl_vowel=ia]
   3   {uu}   pl_vocalism:[measure=(h,ll), pl_vowel=u]

Plural vocalisms are placed on a separate tape. The vocalisms listed above apply to bimoraic stems (H and LL); other stems take the plural vowel melody {ai}. For purposes of clarity, and in order to concentrate on the broken plural problem, the vocalisms {a} and {u} are entered as {aa} and {uu} to avoid writing spreading rules.

### 6.2.2 The Grammar

The following grammar builds on the moraic grammar given in section 4.2.2. The additional rules appear in (62).

(62) BROKEN PLURAL GRAMMAR
   a. CVCC-stem Rules:
      R12   B:Φ:   * - $C_1V_{p1}C_2V_{p2}$ - *   ⇒   * - ($\sigma_{\mu\mu}$,$C_1C_2$,$V_s$,$V_{p1}V_{p2}$+) - ($\sigma$,*, , )
   b. CVCVC-stem Rules:
      R13   B:Φ:   * - $C_1V_{p1}C_2V_{p2}$ - *   ⇒   * - ($\sigma_\mu$,$C_1$,$V_{s1}$,$V_{p1}V_{p2}$+) ($\sigma_\mu$,$C_2$,$V_{s2}$, ) - *
      R14   B:Φ:   * - $C_1V_{p1}C_2V_{p2}$ - *   ⇒   * - ($\sigma_\mu$,$C_1$,V,$V_{p1}V_{p2}$+) ($\sigma_\mu$,$C_2$, , ) - *
   c. CVCVVC-stem Rules:
      R15   B:Φ:   * - $C_1aC_2$aaʔi - *   ⇔   * - ($\sigma_\mu$,$C_1$,$V_1$, ) ($\sigma_{\mu\mu}$,$C_2$,$V_2$, ) - *   [number=pl]
      R16   B:Φ:   * - $C_1aC_2$aaʔi - *   ⇔   * - ($\sigma_\mu$,$C_1$,V, ) ($\sigma_{\mu\mu}$,$C_2$, , ) - *   [number=pl]
   d. CVVCVVC-stem Rules:
      R17   B:Φ:   * - $C_1$awaa - *   ⇔   * - ($\sigma_{\mu\mu}$,C,V, ) - ($\sigma_{\mu\mu}$,*,*, )   [number=pl]
      R18   B/Φ:   * - Cii - *   ⇔   (S,*,*, ) - ($\sigma_{\mu\mu}$,C,V, ) - ($\sigma$,*, , )   [number=pl]
   e. Rare Singular Rules:
      R19   B:Φ:   * - $C_1$awaa - *   ⇔   * - $C_1V_1V_1$ - $C_2V_2$   [number=pl]
      R20   B/Φ:   * - $C_2$i - *   ⇔   $C_1V_1V_1$ - $C_2V_2$ - $C_3$   [number=pl]
   f. Quadrilateral Rules:
      R21   B:Φ:   * - $C_1aC_2$aa - *   ⇔   * - $C_1V_1C_2$ - $C_3V_2$   [number=pl]
      R22   B/Φ:   * - $C_3$i - *   ⇔   $C_1V_1C_2$ - $C_3V_2$ - $C_4$   [number=pl]
   where: Cs=radicals; Vs=vowels; S ∈ {$\sigma_\mu$,$\sigma_{\mu\mu}$}.

Moving from (62f) upward, R21-R22 derive quadrilateral plurals, e.g. /ĵanaadib/, from the a-templatic singular /ĵundub/. R21 parses B:Φ of the singular, i.e. /jun/, and maps it to the iambic plural template in SURF. R22 processes the residue B/Φ, excluding the obligatory final ($\sigma$), i.e. /du/, and maps it to the surface after overwriting the singular vowel with the plural one. The lexical contexts ensure that the proper rule is applied to B:Φ or B/Φ. The derivation of /ĵanaadib/ is illustrated in (63a).

(63) MULTI-TAPE TWO-LEVEL DERIVATION OF /ĵanaadib/ AND /xawaatim/
   a. /ĵanaadib/                          b. /xawaatim/

| ĵun | du | b | + | LT |
|---|---|---|---|---|
| 21 | 22 | 1 | 8 | |
| ĵanaa | di | b | | ST |

| xaa | ta | m | + | LT |
|---|---|---|---|---|
| 19 | 20 | 1 | 8 | |
| xawaa | ti | m | | ST |

The numbers between the two-levels indicate the rules in sections 4.2.2 and in (62) which sanction the moves.

   R19-R20 handle the rare CVVCVC singular forms in a similar manner, expect that the default [w] is inserted in place of the second C of the iambic template in SURF. The rules are illustrated in (63b).

   The remaining rules handle templatic nouns. R17-R18 derive plurals of the form /ĵaamuus/ → /ĵawaamiis/. R17 parses B:Φ (i.e. /ĵaa/) by reading $\sigma_{\mu\mu}$ from the first tape, C from the second and V from the third, i.e. LEX = ($\sigma_{\mu\mu}$,C,V, ); this is exactly the same as LEX in R5 (section 4.2.2), except that the corresponding surface form is the iambic plural template Cawaa, with [w] being inserted for the lack of a second C. RLC = ($\sigma_{\mu\mu}$,*,*,*) ensures that the rule is invoked on the B:Φ portion of the singular stem and not on the residue B/Φ. R18 is similar to R7, but maps B/Φ to Cii on the surface, overwriting the singular vowel with the plural one; the lexical contexts ensure that the rule is only invoked on B/Φ and not on B:Φ. The derivation of /ĵawaamiis/ is illustrated in (64a) (for purposes of comparison, the derivation of the singular is repeated in (64b).

(64) MULTI-TAPE TWO-LEVEL DERIVATION OF /ĵawaamiis/
   a. /ĵawaamiis/ 'plural'                b. /ĵaamuus/ 'singular'

| a | u | | + | VT |
|---|---|---|---|---|
| ĵ | m | s | + | RT |
| $\sigma_{\mu\mu}$ | $\sigma_{\mu\mu}$ | $\sigma$ | + | PT |
| 17 | 18 | 2 | 9 | |
| ĵawaa | mii | s | | ST |

| a | u | | + | VT |
|---|---|---|---|---|
| ĵ | m | s | + | RT |
| $\sigma_{\mu\mu}$ | $\sigma_{\mu\mu}$ | $\sigma$ | + | PT |
| 5 | 7 | 2 | 9 | |
| ĵaa | muu | s | | ST |

Few points are noteworthy: Firstly, R17-R18 are implicitly finding a singular derivation from the lexicon, but mapping such derivation to the corresponding plural form on the surface; note that the vowels which appear on VT in (64a) are those of the singular /ĵaamuus/. Secondly, the rules in (62), with the feature structures [number=pl], are all obligatory. Thirdly, the rule feature structure [number=pl] must unify with the lexical feature structure [number=N] in section 4.2.1. Finally, the lexical structures in (64a-b) are equivalent since the lexical expression in R17-R18 are equivalent to those in R5 and R7, respectively.

R15 handle CVCVVC stems, e.g. /ĵaziir/. Remember that the parsing function $\Phi$ cuts through the second syllable resulting in B:$\Phi$ = /ĵazi/ and B/$\Phi$ = /ir/. In order not to split the second syllable, R15 maps both B:$\Phi$ and B/$\Phi$ to the corresponding plural form /$C_1 aC_2 aa\text{ʔ}i$/. It also inserts the default [w], realised as [ʔ]. R16 operates in a similar fashion, but reads only one singular V; this takes care of singular forms with spreading, e.g. /saħaab/ → /saħaaʔib/. Examples appear in (65o).

(65) MULTI-TAPE TWO-LEVEL DERIVATION OF /ĵazaaʔir/ AND /saħaaʔib/

a. /ĵazaaʔir/

| ai | | + | VT |
|---|---|---|---|
| ĵz | r | + | RT |
| $\sigma_\mu \sigma_{\mu\mu}$ | $\sigma$ | + | PT |
| 15 | 2 | 9 | |
| ĵazaaʔi | r | | ST |

b. /saħaaʔib/

| a | | + | VT |
|---|---|---|---|
| sħ | b | + | RT |
| $\sigma_\mu \sigma_{\mu\mu}$ | $\sigma$ | + | PT |
| 16 | 2 | 9 | |
| saħaaʔi | b | | ST |

R12-R14 handle bimoraic stems: R12 handles CVCC stems, e.g. /nafs/ → /nufuus/, and R13 handles CVCVC stems, e.g. /raĵul/ → /riĵaal/. R14 is equivalent to R13, but processes stems which experience spreading in the singular, e.g. /ʔasad/ → /ʔusuud/. Since bimoraic stems do not take the plural vowel melody {ai}, but a lexically marked one, the proper plural melodies are read from the fourth tape. In other words, the rules read two vowel melodies: singular from tape 2 and plural from tape 4. The former determines if a singular form does exists, and the latter is mapped to the vowels on the surface. The following illustration demonstrate how the rules are invoked.

(66) MULTI-TAPE TWO-LEVEL DERIVATION OF BIMORAIC STEMS

a. CVCC stem

| uu+ | | | $VT_{pl}$ |
|---|---|---|---|
| a | | + | VT |
| nf | s | + | RT |
| $\sigma_{\mu\mu}$ | $\sigma$ | + | PT |
| 12 | 2 | 9 | |
| nufuu | s | | ST |

b. CVCVC stem

| ia+ | | | $VT_{pl}$ |
|---|---|---|---|
| au | | + | VT |
| rĵ | l | + | RT |
| $\sigma_\mu \sigma_\mu$ | $\sigma$ | + | PT |
| 13 | 2 | 9 | |
| riĵaa | l | | ST |

c. CVCVC stem (spreading)

| uu+ | | | $VT_{pl}$ |
|---|---|---|---|
| a | | + | VT |
| ʔs | d | + | RT |
| $\sigma_\mu \sigma_\mu$ | $\sigma$ | + | PT |
| 14 | 2 | 9 | |
| ʔusuu | d | | ST |

Since Vs on the second tape can unify with any vowel in the lexicon, the two-level module produces many analyses depending on the number of vocalisms in the lexicon, but only the ones with the proper singular vocalic melody is parsed successfully by the parser module using the following grammar (in addition to the rules in section 4.2.2).

(67) MORAIC MORPHOTACTIC GRAMMAR
```
noun_stem:[measure=M, number=pl] —→
    pl_vocalism:[measure=M, pl_vowel=PV]
    pattern:[measure=M, number=pl]
    root:[measure=M, sing_vowel=SV, pl_vowel=PL]
    vocalism:[sing_vowel=SV].
```

To summarise: The lexical part builds on the moraic lexicon, adding plural vocalisms which sit on their own lexical tape. For each singular class, there are two rules: the first finds B:$\Phi$ and maps it to the iambic plural template on the surface (it also handles the insertion of [w], realised sometimes as [ʔ], if needed); the second finds B/$\Phi$ and maps it to the surface, but with the proper plural vowel melody. In the case of bimoraic stems, where the plural vowel melody is lexically marked, two vowel melodies are read: a singular one from tape 2 which ensures that a singular form does exist, and a plural one from tape 4 which maps to the vowels on the surface. Since the singular vowel melody does correspond to any vowel on the surface, the two-level module produces many analyses, but only those sanctioned by the morphosyntactic grammar are considered well-formed.

# 7 CONCLUSION

This paper presented three illustrative grammars for handling Arabic non-linear morphology based on CV-, moraic- and affixational models. The following is a summary of the three analyses:

- The CV-based grammar was applied to the verbal stem. It requires that verbal templates are indexed, resulting in a relatively large pattern lexicon (with respect to the other pattern lexica). Because of the nature of the template (i.e. a sequence of Cs and Vs), each surface subsequence sanctioned by a two-level rule consists of only one character.

- The moraic analyses was applied to nominal stems. The pattern lexicon in this case is smaller than the CV one (even if we were dealing with the same data, it would be smaller). A two-level rule in the moraic grammar sanctions a whole syllable (or an extrasyllabic element) which reduces the number of lexical/surface subsequences produced. Reduction in the number of subsequences is proportional to reduction in time complexity.

- The affixational model was applied to the verb. Its pattern lexicon maintains only one template. Affix morphemes are placed in a separate lexicon which is in line with autosegmental theory, i.e. each morpheme in a stem sits on a separate tier.

We conclude that for the analyses of Arabic stems, the most efficient grammars are moraic for templatic stems, and affixational for a-templatic stems.

Finally, this paper discussed the hitherto computationally untreated problem of the broken plural. It was shown that the broken plural can be derived from lexical morphemes (patterns, roots and vocalisms) via the implicit derivation of the singular. Such derivation respects two-level theory.

One aspect of Arabic orthography has not been addressed in this paper: partial vocalisation where any short vowel may be omitted on the surface. This issue has been discussed in Kiraz (1994).

The paper has demonstrated that (multi-tape) two-level morphology is adequate for describing Arabic non-linear morphology.